\documentclass[12pt]{iopart}

\usepackage{iopams}  

\usepackage[utf8]{inputenc} 
\usepackage[T1]{fontenc}
\usepackage{url}
\usepackage{ifthen}
\usepackage{cite}
\expandafter\let\csname equation*\endcsname\relax
\expandafter\let\csname endequation*\endcsname\relax
\usepackage[cmex10]{amsmath}

\usepackage{stmaryrd}
\usepackage{amssymb}
\usepackage{bm}
\usepackage{balance}
\usepackage{hyperref}
\usepackage{xcolor}

\DeclareMathOperator{\Var}{Var}

\usepackage[pdftex]{graphicx}
\DeclareGraphicsExtensions{.pdf,.jpeg,.png}

\begin{document}

\title[Universality of Noiseless Linear Estimation]{On the Universality of Noiseless Linear Estimation\\with Respect to the Measurement Matrix}

\author{Alia Abbara$^1$, Antoine Baker$^1$, Florent Krzakala$^1$ and Lenka Zdeborov\'a$^2$}
\address{$^1$ Laboratoire de Physique de l’Ecole normale sup\'erieure, Universit\'e PSL, CNRS, Sorbonne Universit\'e, Universit\'e Paris-Diderot, Sorbonne Paris Cit\'e, Paris, France \\
                    $^2$ Institut de physique th\'eorique,
                     	 Universit\'e Paris Saclay,
                      CNRS, CEA, 91191 Gif-sur-Yvette, France }

\ead{alia.abbara@ens.fr}

\begin{abstract}
  In a noiseless linear estimation problem, one aims to reconstruct a vector~$\bf x^*$ from the knowledge of its linear projections ${\bf y} = \Phi {\bf x^*}$. There have been many theoretical works concentrating on the case where the matrix $\Phi$ is a random i.i.d. one, but a number of heuristic evidence suggests that many of these results are universal and extend well beyond this restricted case. Here we revisit this problematic through the prism of development of message passing methods, and consider not only the universality of the $\ell_1$ transition, as previously addressed, but also the one of the optimal Bayesian reconstruction. We observed that the universality extends to the Bayes-optimal minimum mean-squared (MMSE) error, and to a range of structured matrices.
\end{abstract}

%
%
%
%
%


\section{Introduction}
The problem of recovering a signal through the knowledge of its linear projections is ubiquitous in modern information theory, statistics and machine learning. In particular, many applications require to reconstruct an unknown $n-$dimensional signal vector ${\bf x^*}$ from the linear projections
\begin{equation}
{\bf y} = \Phi {\bf x^*}\, ,\label{def:problem}
\end{equation}
where ${\bf y}$ is a m-dimensional vector, and $\Phi$ is a $m \times n$ random matrix. For instance, if $ {\bf x^*}$ is sparse, this task of estimating the signal from its linear {\it random} projections is at the roots of compressed sensing \cite{candes2006near}. A fundamental question in the field is how much the algorithmic and the information theoretic performance depends on the choice of the random matrix $\Phi$. 

In the present letter, we concentrate on the noiseless and asymptotic, large $n$, regime with a fixed value $\alpha\!=\!m/n$. We consider $\bf x^*$ to be $k$-sparse, i.e. to have only $k$ non-zero values, and we shall work in the limit where $n\to \infty$, $k\to \infty$, and a finite value of $\rho\!=\!k/n$. In such setting, a classical result is the following: for random matrices $\Phi$ with independent standard Gaussian entries, the (convex) reconstruction with $\ell_1$ penalty displays a precisely determined phase transition. For a certain region in the ($\alpha,\rho$)-phase diagram, it typically finds back the vector~$\bf x^*$, being the sparsest solution, whereas outside that region, it typically fails. The obundary between these two regions is called the Donoho-Tanner line~\cite{donoho2005sparse}. It has been shown empirically that the very same phase transition location seems to hold for a wider range of random matrix ensembles, see e.g. \cite{donoho2009observed,Monajemi1181}, suggesting a large universality of the Donoho-Tanner phase transitions. {Another line of work showed that the convex $\ell_1$ reconstruction problem can be treated through conic geometry, and the success probability of signal recovery only depends on a geometric number characterizing a subcone (statistical dimension or Gaussian width)~\cite{chandrasekaran2010,amelunxen2014}.}

Here we investigate the universality of the phase transition not only for the $\ell_1$ transition, but also to the performance of the optimal Bayesian reconstruction. We analyze this question  through the prism of information theory, message passing methods, and random matrix theory. We shall see that the universality indeed extends to a more generic set of properties than the $\ell_1$ transition, such as the minimum mean-squared (MMSE) error or the easy-hard phase transition for optimal Bayesian learning, and empirically to structured matrices such as the one appearing in \cite{pennington2017nonlinear,liao2018spectrum}.

We note that investigation of universality are very common to physics problems, and understanding how large is the class of model for which a given result applied is a very fundamental question. The message-passing-based algorithm that we investigate in this paper to demonstrate the universality also has their origin in pysics works, such as \cite{thouless1977solution}.

\section{A short review of results for i.i.d. random matrices}
\label{sec:iid-matrix}
A first well-understood case of universality holds for random matrices $\Phi$ where all the elements are generated i.i.d. from a well-behaved distribution -with zero mean and unit variance- which all exhibit the same transitions as Gaussian random matrices. This is known for multiple retrieval problems:
\subsection{$\ell_1$ recovery}
Consider for instance the Donoho-Tanner line\cite{donoho2005sparse} that regulates the $\ell_1$ recovery. Thanks to the approximate message passing solver (see below) that has been shown to be universal with respect to all i.i.d. distributions with finite moments \cite{bayati2011dynamics,bayati2015universality}, we know that the Donoho-Tanner phase transition is the same for all such random matrices. 

\subsection{Information theoretic optimal reconstruction}
There has been a considerable amount of work in the information theory community on the computation of the mutual information and on the MMSE for problems such as (\ref{def:problem}) with Gaussian matrices. In particular, following the replica method from statistical physics  (the Tanaka formula \cite{tanaka2001statistical}), a heuristic formula has been postulated in different situations, see e.g. \cite{tulino2013support,krzakala2012probabilistic,krzakala2012statistical,zhu2013performance}. This heuristic replica result has been recently rigorously proven in a series of papers \cite{barbier2016mutual,barbier2016mutual,reeves2016replica}. In a more recent proof \cite{barbier2017phase}, it has been shown, again, that the formula is not specific to Gaussian i.i.d. matrices, but that any matrix with i.i.d elements of unit variance and zero mean leads to the same exact result for the mutual information and the MMSE.

\subsection{Hard phase for Bayesian decoders}
A third interesting point is to ask about tractacle decoders that aim to perform the optimal Bayesian estimation, i.e. with a perfect prior knowledge on the distribution of  $\bf x^*$. For simplicity, consider for instance the case where each element of $\bf x^*$ has been sampled from a Gauss-Bernoulli distribution:
$$x_i \sim (1-\rho) \delta(x) + \rho {\cal N}(0,1)\, .$$ 
In this case, the best known solver is again AMP, using a Bayesian decoder (instead of the soft thresholding function for $\ell_1$ recovery) \cite{vila2011expectation,montanari2012graphical,krzakala2012probabilistic,krzakala2012statistical}. Interestingly, it shares with the $\ell_1$ recovery a similar phase transition: for a certain region in the ($\alpha,\rho$) plane it typically finds back the vector $\bf x^*$, whereas outside that region it fails. We shall denote the limit between these regions the "Bayesian hard-phase" transition. The "Bayesian hard-phase" line, that has been precisely computed in \cite{krzakala2012probabilistic,krzakala2012statistical} is always better than the Donoho-Tanner line (as it should, since it exploits additional information).  Once more, the universality of AMP shows that this phase transition is not restricted to Gaussian matrices, but extends as well to all (well normalized) i.i.d. matrices.

The fact that these three properties (the $\ell_1$, the hard-phase line, as well as the MMSE) are universal for all i.i.d. matrices makes the case for Gaussian computations, as done in theoretical computation, stronger. We shall see that this universality extends well beyond these simple cases.

\section{Random rotationally invariant matrices}
\label{sec:rot-matrix}
Moving away from the well-known i.i.d. examples, we start by considering a much larger set of random matrices defined through their  singular value decomposition (SVD): any real matrix $\Phi$ can be decomposed into ${\Phi = U \Sigma V}$, with $U$ and $V$ orthogonal matrices, and $\Sigma$'s elements being $\Phi$'s singular values. We shall look at the left rotationally invariant random matrix ensemble: these are matrices $\Phi$ that can be written as
$$\Phi = U \Sigma V$$
with an arbitrary rotation matrix $U$ and singular values $\Sigma$, but where the matrix $V$ has been randomly (and independently of $\Sigma$ and $U$) generated from the Haar measure (that is, uniformly from all possible rotations). 

When the singular values are different from zero, it is straightforward to justify the universality property for matrices from this subclass. We start by the definition of the problem: we wish to find $\boldsymbol{x}$ such that
\begin{equation}
\boldsymbol{y}=\Phi \boldsymbol{x}=U\Sigma V \boldsymbol{x}.
\label{def_problem}
\end{equation}
If $m \le n$, then $\Sigma$ is written as $\Sigma=
\left[
\begin{array}{c|c}
\tilde{\Sigma} & 0 \\
\end{array}
\right]$  and we define 
$$\Sigma^{inv}=
\left[
\begin{array}{c}
\tilde{\Sigma}^{-1}\\
\hline
0
\end{array}
\right]  \text{ such that  } \Sigma^{inv} \Sigma=
\left[
\begin{array}{c|c}
I_m & 0\\
\hline
0 & 0
\end{array}
\right]. $$
Multiplying~\eqref{def_problem} on both sides by $U^T$, and then by $\Sigma^{inv}$; one reaches
\begin{equation}
\boldsymbol{\tilde y} = \Sigma^{inv} U^T \boldsymbol{y} = \tilde{V} \boldsymbol{x}
\label{eq:magic1}
\end{equation}
where $\tilde{V}$ is a $m\times n$ matrix composed of the first $m$ lines of $V$. If instead $m > n$, $\Sigma$ is written as $$\Sigma^{inv}=
\left[
\begin{array}{c}
\tilde{\Sigma}\\
\hline
0
\end{array}
\right]
$$
and we define $\Sigma^{inv}= \left[
\begin{array}{c|c}
\tilde{\Sigma}^{-1} & 0 \\
\end{array}
\right]$ such that $\Sigma^{inv} \Sigma = I_n$. Multiplying~\eqref{def_problem} by $U^T$ then $\Sigma^{inv}$, we obtain
\begin{equation}
\boldsymbol{\tilde y} = \Sigma^{inv} U^T \boldsymbol{y} = V \boldsymbol{x}.
\end{equation}
In both cases, we thus see that the problem has been transformed ---in a constructive way--- into a standard linear system with the sensing matrix $\tilde{V}$ when $m \le n$ being a (sub-sampled) random rotation one, or sensing matrix $V$ when $m > n$. This shows that all rotationally invariant matrices, {which satisfy $U$ and $\Sigma$'s independence on $V$, can be transformed the same way and are in the same universality class as far as noiseless linear recovery is concerned, i.e. they will display the same phase transitions.}

Since Gaussian i.i.d. matrices belong among random rotationally invariant matrices (in this case $\Sigma$ follows the Marcenko-Pastur law \cite{tulino2004random}) this means that all the information theoretic rigorous results {(such as phase transitions and MMSE value)} with zero noise for random Gaussian matrices applies verbatim to all rotationally invariant ensemble, {as long as the SVD's matrices $U$ and $\Sigma$ are independent of $V$.} This is a very strong universality, that applies to the three cases (1, 2, 3) from sec.~\ref{sec:iid-matrix}. Note that the universality of the Donoho-Tanner line with rotationally invariant matrices was already hinted by the replica method \cite{kabashima2009typical}. 

Notice, however, that the above construction depends crucially on the fact that we consider here noiseless measurements. It would not work if an additional Gaussian noise were added in eq.~(\ref{def:problem}): in this case, the transformation would make the i.i.d. Gaussian noise a correlated one. Indeed, the replica formula for noisy measurements underlines that the MMSE depends on the precise set of matrices in noisy reconstruction \cite{takeda2006analysis,tulino2013support} (this formula is not yet fully rigorous, but see \cite{barbier2018mutual} for a proof in a restricted setting). Any differences, however, must go to zero in the noiseless limit. 

\section{Approximate Message Passing}
Having discussed the universality with respect to random rotationally invariant matrices, we now wish to discuss its effect on specific solvers, concretely the message passing algorithms.
\subsection{AMP} We first consider the original approximate message passing (AMP)~\cite{donoho2009message} to compute the phase transition between the phase where the algorithm reconstructs $\bf x^*$ perfectly, and the one where reconstruction may be possible but is not achieved by the algorithm.  AMP is an iterative algorihm that follows:
\begin{align*}
&{\bf \hat x}^{t+1}=\eta_t(\Phi^T {\bf \hat x}^t)\\
&{\bf z}^t={\bf y}-\Phi{\bf \hat x}^t + \frac{1}{\alpha}{\bf z}^{t-1}\langle \eta_{t-1}'(\Phi^T {\bf z}^{t-1}+{\bf \hat x}^{t-1})\rangle.
\end{align*}
where $t$ is the iteration index, ${\bf x}^t$ is the current estimate of ${\bf x^*}$, ${\bf z}^t$ the current residual,  $\langle \cdot \rangle$ is an averaged sum of components, and $\eta_t$ is a prior-dependent threshold function applied component-wise (the soft thresholding for $\ell_1$, or the Bayesian decoder \cite{krzakala2012probabilistic,krzakala2012statistical}). 

One of the most interesting features of AMP is that, if $\Phi$ is a Gaussian i.i.d. matrix, its mean squared error (MSE) $\sigma_t$ can be tracked accurately by the state evolution formalism~\cite{donoho2009message,bayati2011dynamics,bayati2015universality}. State evolution is a relatively simple recursive equation:
\begin{equation}
\sigma_{t+1}^2 = \Psi(\sigma_{t}^2)\, ,~~
\Psi(\sigma^2)={\mathbb E}\left[ \left(
\eta_t(X+\frac{\sigma}{\sqrt{\alpha}}Z)-X
\right)^2  \right] \, ,
\end{equation}
where the expectation is with respect to independent random variables $Z \sim {\cal N}(0, 1)$ and $X$, whose distribution coincides with the empirical distribution of the entries of $x^*$. Analyzing the evolution of this equation for the $\ell_1$ decoder yields the Donoho-Tanner line\cite{donoho2009message}, while using the Bayesian decoder it yields the hard-phase line for Bayesian decoding~\cite{krzakala2012probabilistic}.

It would be interesting to use AMP for rotationally invariant matrices. In order to do this, we follow the construction of sec.~\ref{sec:rot-matrix}: starting from equation (\ref{eq:magic1}) we then multiply by $\Sigma_0$, a $m \times m$ diagonal matrix with singular values sampled from Marcenko-Pastur law (singular values of a Gaussian i.i.d. matrix \footnote{The singular values of a Gaussian matrix are correlated, so in fact we may want to  generate $\Sigma_0$ by first generating a random Gaussian matrix, and then calculating its singular values.}),
and $U_0$ a $m \times m$ Haar-generated orthogonal matrix, {thus ensuring that $\Sigma_0$ and $U_0$ are generated independently of $V$}:
\begin{align}
U_0 \Sigma_0 \tilde{\Sigma}^{-1} U^T \boldsymbol{y} &= U_0 \Sigma_0 \tilde{V} \boldsymbol{x}\\
\boldsymbol{y'} &= \Phi' \boldsymbol{x}.\label{eq:trick}
\end{align}
After this transformation,  $\Phi'=U_0 \Sigma_0 \tilde{V}$ is a random matrix that belongs to an ensemble very close to the Gaussian i.i.d. matrices ensemble. {In fact, a recent work showed that AMP applied to a Gaussian matrix follows the same state evolution as matrices such as $\Phi'$ where $U_0, \tilde{V}$ are uniform orthogonal matrices and $\Sigma_0$ diagonal's elements are singular values sampled from the Marcenko-Pastur law \cite{takeuchi2019unified}.} Combining this result with the matrix transformation, we have thus constructively mapped the noiseless reconstruction problem back to the well-understood noiseless compressed sensing case for a Gaussian i.i.d. matrix, where we can safely apply the algorithm, and its state evolution. 
In the section \ref{sec:numerics}, we apply this matrix transformation for numerical experiments using AMP.
\subsection{Vector-AMP}
While the transformation trick allows to make AMP work with random rotationally invariant matrices, another alternative is to work directly with a dedicated solver. To this means, different but related approaches
were proposed \cite{takeda2006analysis,SAMP}, in particular, using the
general expectation-propagation (EP) \cite{Minka,EP} scheme.  Ma and Ping proposed a variation of EP called OAMP \cite{ma2017orthogonal} specially adapted
to rotation matrices. Rangan, Schniter and Fletcher introduced a similar approach called VAMP \cite{VAMP} and proved that it follows state evolution equations corresponding to the
fixed point of the replica potential \cite{takeda2006analysis,tulino2013support,barbier2018mutual}. The multi-layer AMP algorithm of\cite{Manoeletal} also display the same fixed point. 

We shall concentrate here on the VAMP (Vector-AMP) approach, and for a moment, put back a small additional random Gaussian i.i.d. noise of variance $\Delta$ in the measurement in eq.~(\ref{def:problem}) as it is needed for stating the algorithm. VAMP then consists in the following fixed-point iteration:
\begin{equation}
    \begin{aligned}
        &\bm{u}_{\ell}^{t + 1} = \frac{
        \bm{\hat x_l}^t}{\langle \Var_\ell^t (\bm{x}) \rangle} -
        \bm{u}_{r}^t, \qquad
        \rho_{\ell}^{t + 1} = \frac{1}{\langle \Var_\ell^t
        (\bm{x}) \rangle} - \rho_r^t, \\
        &\bm{u}_r^{t + 1} = \frac{
        \bm{\hat x_r}^t}{\langle\Var_r (\bm{x}) \rangle} -
        \bm{u}_\ell^t, \qquad
        \rho_r^{t + 1} = \frac{1}{\langle \Var_r^t
        (\bm{x}) \rangle} - \rho_\ell^t,
    \end{aligned}
\end{equation}
where we denote by $\mathbb{E}_{\ell, r}^t$ the expectation w.r.t. the tilted distributions ${\tilde{Q}_{\ell, r}^t (\bm{x}) \propto P_{\ell, r}
(\bm{x}) Q_{\ell, r}^t (\bm{x})}$, 
and  by $\Var_{r, \ell}^t (\bm{x})$ the variance
of these distributions. Here, we have defined 
 $Q_{l,r}({\bf x})=e^{-\frac 12 \rho_{l,r} \bm{x}^T\bm{x} + \bm{u}_{l,t}^T \bm{x}}$, $P_l(\bm{x}) \propto e^{-||{\bf y}-\Phi {\bf x}||_2^2/2\Delta}$ and $P_r(\bm{x})$ is the prior used in the algorithm (i.e. the Laplace prior for the $\ell_1$ model, or the actual distribution of the signal for Bayesian reconstruction). In particular
\begin{equation}
    \begin{aligned}
\bm{\hat x_l}^t &= (\bm{\Phi}^T \bm{\Phi} + \Delta \rho_r^t I_p)^{-1} (\bm{\Phi}^T \bm{y} +  \Delta\bm{u}_r^t),\\
    \langle \Var_\ell^t (\bm{x}) \rangle &= \frac{ \Delta}{N} \rm{Tr} (\bm{\phi}^T \bm{\Phi} +  \Delta \rho_r^t I_p)^{-1},
        \end{aligned}
\end{equation}
where, as for AMP, we define the denoiser that yields the estimates of $x$ 
 by $z (u, \rho) = \int dx P_r(x) e^{-\frac12 \rho x^2 + u x}$,
\begin{equation}
\begin{aligned}
    {(\hat x_r)}_j &= \frac{\partial}{\partial u} \log z (u, \rho)
        \Big|_{u_{\ell k}^t, \rho_\ell^t},\\
    \langle \Var_r^t (\bm{x}) \rangle &= \frac{1}{n} \sum_{j = 1}^p
        \frac{\partial^2}{\partial u^2} \log z (u, \rho) \Big|_{u_{\ell k}^t,
        \rho_\ell^t}.
    \label{eq:ec_avgr}
    \end{aligned}
\end{equation}
Again, the performance of the recursion can be analyzed rigorously through the state evolution \cite{VAMP}. For simplicity, let us concentrate on the Bayes optimal case in which case the state evolution can be closed on the variables (see \cite{VAMP}):
\begin{equation}
\sigma^t  = \langle \Var_r^t (\bm{x}) \rangle~\rm{and}~
\epsilon^t  = \langle \Var_l^t (\bm{x}) \rangle \, ,
\end{equation}
by writing 
\begin{equation}
\begin{aligned}
\sigma^t (\rho^t_l) &= \Psi((\rho^t_l)^{-1}) \\
\epsilon (\rho^t_r) &=  \Delta {\mathbb{E}}\left[ \frac 1{\Sigma^2+ \Delta \rho_r^t}\right] =  \Delta S_{\Sigma^2}(-\Delta \rho^t_R)
    \end{aligned}
\end{equation}
where the expectation is above the distribution of the singular values $\Sigma$ of the matrix $\Phi$, and where we recognize the Stieljes transform $S_{X}(r)={\mathbb{E}}\left[ 1/{X - r}\right] $. 

Though this transform, we see that the performance depends crucially on the distribution of eigenvalues. Let us now go back on the noiseless limit when $\Delta \to 0$ and analyze how the universality shows up. Consider again the Stieltjes transform: out of the $n$ singular values of the $n \times n$ matrix $\bm{\Phi}^T \bm{\Phi}$, we shall have $(1-\alpha)n$ of them to be zero (assuming $\alpha<1$) while the rest are positive (since $m<n$). In this case, the limit $r \to 0$ of the Stieltjes transform will behave as $S_{X}(r) \approx-(1-\alpha)/r$ so that 
$$
\lim_{\Delta \to 0} \epsilon (\rho^t_r) = \frac{1-\alpha}{\rho_r^t}\, .
$$
Again, we see that all the complicated dependence on the spectrum of the matrix $\Phi$ has been eliminated. This is a direct, alternative, proof that VAMP will also yield universal results in the zero noise limit for the Bayesian reconstruction. Given that VAMP has the same fixed point as the replica mutual information \cite{tulino2013support,barbier2018mutual}, this argument applies to the replica prediction for the MMSE as well.
\section{Structured matrices}
We now move to very structured matrices, in order to test the universality as well as the quality and the prediction of the state evolution out of its comfort zone. In order to do so, we have considered different matrix ensembles:

 \begin{figure}[htbp]
   \centering
   \includegraphics[width=0.75\textwidth]{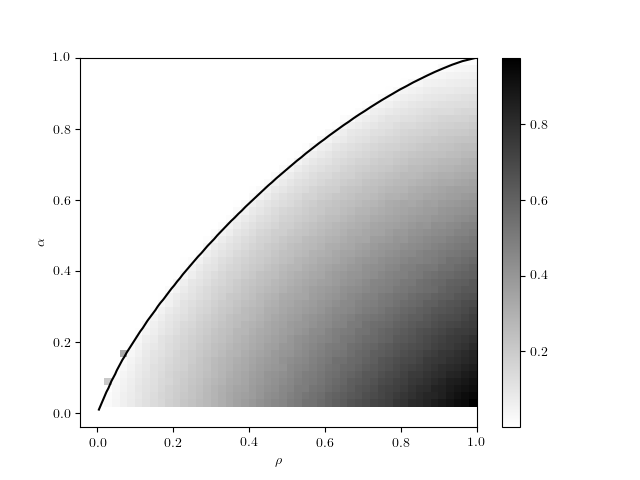}
   \caption{Phase diagram for a DCT matrix (width $n=1000$) in the Bayes-optimal case. The averaged MSE on 50 executions of VAMP is represented by a color-code, displaying a phase transition that matches the theoretical Bayes line for Gaussian i.i.d. matrices (black line). Some finite-size effects can be seen.}
   \label{fig:sim1}
 \end{figure}
 
\begin{figure}[htbp]
   \centering
   \includegraphics[width=0.75\textwidth]{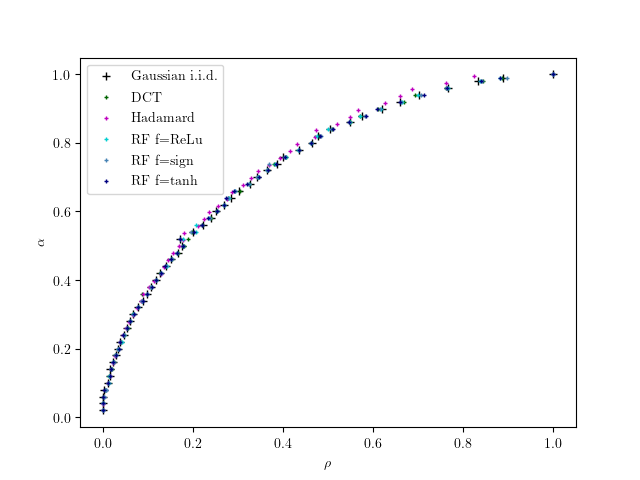}
   \caption{Phase diagram in the $\ell_1$ reconstruction case obtained by averaging on 20 to 50 executions on VAMP. The dots indicate the phase transitions for Gaussian i.i.d., DCT (width $n=2000$), Hadamard matrices ($n=4096$), and random feature  matrices $\Phi=f(WX)$ with ${f = \text{ReLu}}$, ${f = \text{sign}}$, ${f= \tanh}$ ($W$ and $X$ are Gaussian i.i.d. of size $\alpha n \times n$ and $n \times n$ with $n=2000$). They match the theoretical Donoho-Tanner transition for Gaussian i.i.d. matrices (black line).}
   \label{fig:sim2}
 \end{figure}

\subsection{Tested ensembled of matrices}

\paragraph{Discrete cosine transform matrices}
The first ensemble we consider consists in Fourier-like matrices. 
A $n \times n$ discrete cosine transform (DCT) matrix $Y$ is defined by:
\begin{equation}
Y_{jk}=\sqrt{\dfrac{2}{n}}\epsilon_k \cos\left( \dfrac{\pi (2j+1) k}{2n}\right)\, ,
\end{equation}
where $j,k \in \llbracket 0, n-1 \rrbracket$, 
$\epsilon_0=1/ \sqrt{2}$, 
$\epsilon_i=1$ for $i=1,...,n-1$.
We used a sub-sampled version of these matrices in which we picked some rows randomly.

\paragraph{Hadamard matrices}
A natural variant of DCT is given by the Hadamard matrices. $H$ is a $n \times n$ Hadamard matrix if its entries are $\pm 1$ and its rows are pairwise orthogonal, i.e. $H H^T= n I_n$. For every integer $k$, there exists a Hadamard matrix $H_k$ of size $2^k$. These can be created with Sylvester's construction: Let $H$ be a Hadamard matrix of order $n$. Then the partitioned matrix
$${\begin{bmatrix}H&H\\H&-H\end{bmatrix}}$$
is a Hadamard matrix of order $2n$. 

%
\paragraph{Random features maps}
Finally, we wanted to consider here random features maps (RFM) as encountered in nonlinear regression problems. In such settings, a random features matrix $\Phi = f(WX)$ is obtained from the raw data matrix $X$ by means of a random projection matrix $W$ and a pointwise nonlinear activation $f$. Kernel regression models, nonlinear in the original data~$X$, can then be approximately but efficiently solved by the linear estimation problem (\ref{def:problem}), with an appropriate choice for $f$ and the $W$-distribution \cite{rahimi2008random}.
Such matrices, that can be seen as the ouput of a neuron with random weights, have been investigated in particular in the context of neural networks \cite{pennington2017nonlinear,liao2018spectrum}. Indeed, in neural networks configurations with random weights play an important role as they define the initial loss landscape. They are also fundamental in the random kitchen sinks algorithm in machine learning \cite{rahimi2008random} and it is thus of interest to test our understanding of linear reconstructions with AMP and VAMP in this case.

In what follows we will test random features matrices where both $W$ and $X$ are random Gaussian i.i.d. matrices. 

\subsection{Numerical results}
\label{sec:numerics}
\begin{figure}[htbp]
   \centering
   \includegraphics[width=0.75\textwidth]{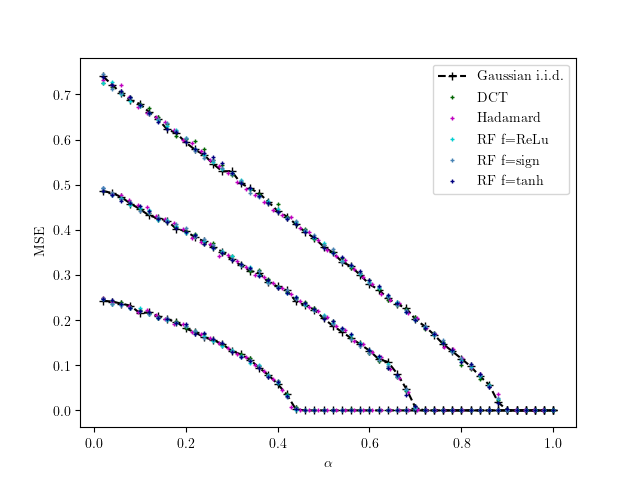}
   \caption{Mean-squared error for $\rho =0.25$, $0.5$ and $0.75$ (bottom to up curves) in the Bayes-optimal case averaged on 20 executions of VAMP for Gaussian i.i.d, DCT, Hadamard, random features matrices $\Phi = f(WX)$ with ${f = \text{ReLu}}$, ${f = \text{sign}}$, ${f= \tanh}$ ($W$ and $X$ are Gaussian i.i.d of size $\alpha n \times n$ and $n \times n$) . The width is $n=2000$ for all matrices.}
   \label{fig:sim3}
 \end{figure}

\begin{figure}[htbp]
   \centering
   \includegraphics[width=0.75\textwidth]{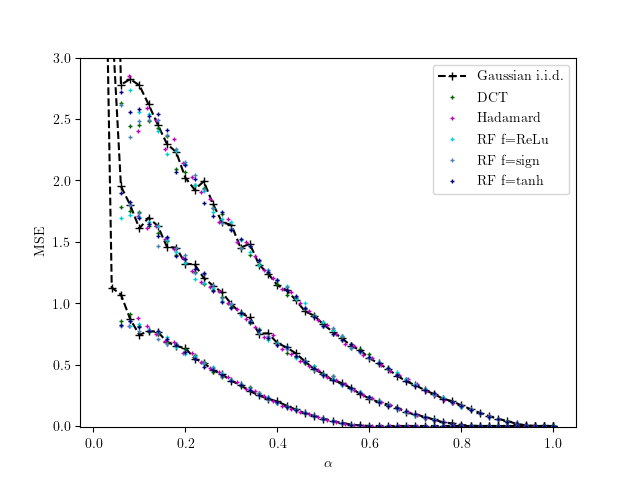}
   \caption{Mean-squared error for $\rho =0.25$, $0.5$ and $0.75$ (bottom to up curves) in the $\ell_1$ reconstruction case averaged on 20 executions of VAMP for Gaussian i.i.d, DCT, Hadamard, random features matrices $\Phi = f(WX)$  with ${f = \text{ReLu}}$, ${f = \text{sign}}$, ${f= \tanh}$ ($W$ and $X$ are Gaussian i.i.d of size $\alpha n \times n$ and $n \times n$). The width is $n=2000$ for all matrices.}
   \label{fig:sim4}
 \end{figure}

We provide the codes used to generate the data on github in the repo
\href{https://github.com/sphinxteam/Universality-CS-2019}{http://sphinxteam/Universality-CS-2019}. To generate Figure \ref{fig:sim1} and \ref{fig:sim2}, we ran VAMP $50$ times on $50\times50$ points spanning the ($\alpha$,$\rho$)-space, and computed the average mean-squared error (MSE) between the signal ${\bf x^*}$ and the reconstructed configuration {\bf x}. The MSE is represented with a color bar (white means perfect reconstruction). For a DCT and a Hadamard matrix, we observe a phase transition in the Bayes-optimal case that matches the theoretical transition for Gaussian i.i.d. matrices. We also ran VAMP for the $\ell_1$ reconstruction problem. Averaging on 20 executions (or 50 for small $\alpha$ where finite-size effects are more important), we recover again a phase transition matching the theoretical Donoho-Tanner line for Gaussian i.i.d. matrices \cite{donoho2009observed}. Besides, we compared the MSE obtained by VAMP at each point of the phase diagram for different matrices. In figures \ref{fig:sim3} and \ref{fig:sim4}, we plot the MSE averaged on 20 executions for $\rho$ fixed  and $\alpha$ ranging between 0 and 1. We get the same error in reconstruction for all matrices, following the MSE for Gaussian i.i.d. matrix for $\rho =0.25$, $0.5$ and $0.75$. We also checked that AMP, provided one uses the trick eq.~(\ref{eq:trick}), reproduce these results as well: indeed the two algorithms returned extremely similar results.

\subsection{Discussion}

Figures of the previous section perfectly illustrate our main point: the universality in noiseless compressed sensing is not limited to the $\ell_1$-type reconstruction as in \cite{donoho2009observed,Monajemi1181}, but extends to other quantities and estimators, such as the hard-phase line in Bayesian reconstruction, and the MMSE. Besides, it is not limited to random orthogonal matrices, but empirically extends to Fourier-type matrices and to the random features maps currently studied in machine learning. It is an open question to extend the proof of state evolution to these challenging matrices. It would be interesting to find a good creterion to identify which matrices satisfy this universality and which do not; this is something that we are yet unable to predict in advance. An example of structured matrices that do not seem to follow these universal phase transitions is given by Haar wavelet matrices, which can be defined recursively by:
$$W_2={\begin{bmatrix}1&1\\1&-1\end{bmatrix}}\text{ and } W_{2k}= {\begin{bmatrix}H_k \otimes [1,-1]\\I_k \otimes [1,1]\end{bmatrix}} $$
where $I_k$ is the identity matrix of size $k$ and $\otimes$ is the Kronecker product. In fact, VAMP even fails to converge for these matrices. Investigating this behavior is an interesting direction of research.

\section*{Acknowledgment}
We thank Andre Manoel and Galen Reeves for useful discussions. We acknowledge funding from the ERC under the European Union’s Horizon 2020 Research and Innovation Program Grant Agreement 714608-SMiLe; and from the French National Research Agency (ANR) grant PAIL.

\vspace{1cm}

\bibliographystyle{IEEEtran}
\bibliography{refs}

\end{document}